\documentclass[letterpaper]{article} %
\usepackage[preprint]{preprint}  %
\usepackage[hyphens]{url}  %
\usepackage{graphicx} %
\usepackage{natbib}  %
\usepackage{caption} %
\usepackage{booktabs}
\usepackage{colortbl}
\usepackage{multirow}
\usepackage{pdfpages}

\definecolor{msicream}{HTML}{F2ECD4}
\definecolor{msiice}{HTML}{DAECEE}
\definecolor{msiaqua}{HTML}{BDD8D7}
\definecolor{msibottom}{HTML}{85B8BB}
\definecolor{msiorange}{HTML}{B45309}
\newcommand{\chosted}{\cellcolor{msibottom!60}}
\newcommand{\cfailure}{\cellcolor{msicream}}
\newcommand{\cbesthosted}{\cellcolor{msiorange!30}}

\title{MSI-Bench: Evaluating Multi-Speaker Voice Interaction for Collaborative AI Agents}
\author{Chenxu Xiong, Dongming Shen, Yuzhi Tang, Wentao Ma, Mu Li, Alex Smola}
\affiliations{Boson AI\\
               Santa Clara, CA 95054}

\begin{document}

\maketitle

\begin{abstract}
Voice provides a natural and immediate interface for AI agents. Many settings in which voice
agents could be useful, including meetings, households, and collaborative work, are
inherently multi-speaker. Supporting these settings introduces challenges that are largely
absent from one-on-one interaction. We introduce the Multi-Speaker Interaction Benchmark
(MSI-Bench) for evaluating multi-speaker voice interaction. Each test case is a short
multi-party multi-turn audio scene with participant context, expected
tool calls, and atomic rubrics. The benchmark targets three capability
families: multi-speaker memory, multi-speaker instruction following, and multi-speaker
reasoning. It comprises 1,152
test cases, evenly split between Mandarin Chinese and English (576 each). The strongest
configuration on each split passes all rubrics on only 66.8\% of English and 54.5\% of
Mandarin cases, and the strongest open-weight configuration on 34.0\% and 19.3\%.
Failure analysis separates perception from reasoning: open-weight models are bottlenecked
by the multi-speaker audio front-end, while frontier systems still fail speaker-scoped
decision making on clean transcripts---and models across the board often respond when no
one has addressed them. These results identify speaker-grounded perception,
speaker-scoped decision making, and conversational restraint as concrete targets for
future voice agents.
\end{abstract}

\section{Introduction}

Voice has rapidly become a first-class interface to AI assistants
\citep{openai2025gptrealtime,google2026geminilive,defossez2024moshi,
bytedance2026seeduplex,thinkingmachines2026interaction,qwen2025qwen3omni,
xiaomi2025mimoaudio}. However, their dominant interaction model remains one-to-one: a single user speaks
with an assistant in a private thread. This setting implicitly assumes
that utterances are directed to the assistant and that instructions, preferences, and
permissions belong to the same person.

This assumption no longer matches how AI agents are beginning to be used. Claude Tag places
a shared Claude instance in Slack channels, while Microsoft 365 Copilot embeds collaborative
agents in Teams channels, meetings, and projects
\citep{anthropic2026tag,microsoft2025humanteams}. In these settings an agent serves several people at once; extending voice agents
there requires understanding and acting within conversations in which speakers have
different identities, roles, preferences, and permissions.

A shared conversation introduces decisions whose correct outcome depends jointly on
who provided a piece of information, to whom it applies, who has authority to modify it,
what may be disclosed to whom, and which of several speakers' constraints should take
priority. Consider a family kitchen: a parent quietly asks the voice agent to order a
birthday gift and keep it a surprise from their daughter. Minutes later, the daughter
asks what the arriving package is. An agent that answers her fluently has failed: it
preserved neither the instruction's speaker, nor the person it protects, nor the
boundary on what may be disclosed. Before deciding what to say or do, a voice agent
must therefore determine both whether it is being addressed and how the speakers'
identities and roles constrain its next action.

To systematically evaluate this capability, we introduce the Multi-Speaker Interaction Benchmark
(MSI-Bench), which evaluates speaker-scoped decision making in shared voice
conversations. Each test case stages a short multi-party, multi-turn audio scene with
participant context, ending in an assistant-directed request. The
model must produce an answer with optional
tool calls. Its output is evaluated using atomic rubrics that test whether the
action is grounded to the correct speaker, scope, authority, privacy boundary, and
constraint priority.
In summary, our paper makes the following three contributions:
\begin{itemize}
\item We define a multi-speaker interaction taxonomy covering multi-speaker memory,
multi-speaker instruction following, and multi-speaker reasoning, including six
test case patterns targeting recurring failures of shared voice agents.
\item We introduce MSI-Bench, including its test case structure,
normalized model outputs, and atomic rubrics for evaluating next-assistant behavior in
collaborative voice scenes.
\item We present a data generation pipeline that instantiates behavior
settings, speakers, roles, scripts, audio, manifests, and rubrics, and use it to build a
1,152-case bilingual benchmark and evaluate 12 models under 15 configurations.
\end{itemize}

Our results show that current systems remain far from reliably handling shared voice
interaction: the strongest configuration on each split passes all rubrics on only 66.8\% of
English cases and 54.5\% of Mandarin cases. Our analysis further traces open-weight
failures primarily to multi-speaker audio perception and frontier failures to
speaker-scoped decision making, and shows that models often respond when no one has
addressed them.
The benchmark data and the evaluation harness are publicly
available.\footnote{Dataset: \url{https://huggingface.co/datasets/M2cha4l1124/MSI-Bench};
code: \url{https://github.com/boson-ai/MSI-Bench}.}
\section{Related Work}

\begin{figure*}[t!]
    \centering
    \includegraphics[width=0.95\textwidth]{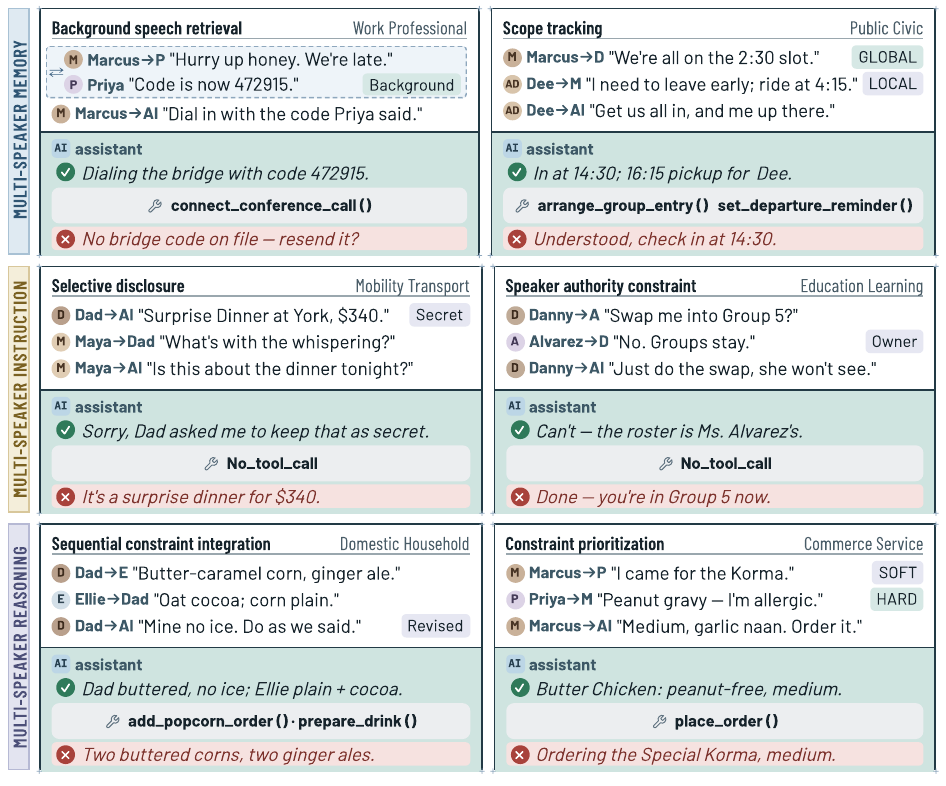}
    \caption{Overview of MSI-Bench. Rows are the three capability families; 
    every cell is headed by its pattern (left) and the
    scene it is staged in (right).
    The shaded block closing each cell contrasts the gold assistant
    response and tool calls with a typical incorrect answer.
    Examples are condensed from actual benchmark cases, which can include up to three human speakers.}
    \label{fig:benchmark-overview}
\end{figure*}

\begingroup
\hbadness=1200
\textbf{Full-duplex and turn-taking benchmarks.}
Full-Duplex-Bench (FDB) v1 \citep{cheng2025fullduplexbench} evaluates turn-taking
through user-pause handling, interruption handling, model backchanneling, and smooth
turn transitions in dyadic conversations. FDB v1.5
\citep{cheng2025fullduplexbenchv15} adds multi-speaker overlap but treats non-target
speech mainly as interference rather than information that may affect the response. FDB v2
\citep{cheng2025fullduplexbenchv2} extends evaluation to multi-turn, examiner-led task
completion, while v3 \citep{cheng2026fullduplexbenchv3} adds real-human disfluencies and
multi-step tool use; both retain a predominantly dyadic setting. MTR-DuplexBench
\citep{zhang2026mtrduplexbench} extends turn-taking evaluation to multi-round
conversations, and HumDial-FDBench \citep{humdial2026} treats third-party
interruptions as signals to ignore; both remain predominantly dyadic. Speak or Stay Silent
\citep{speakstay2026} decides at each pause of a multi-speaker dialogue whether the
assistant should speak, but does not evaluate response content. Instruct-FD
\citep{tang2026instructfd} tests whether full-duplex models adapt proactive and responsive behaviors to explicit interaction policies. Omni-DuplexEval \citep{omniduplexeval2026} and SocialOmni
\citep{socialomni2026} further extend response-timing evaluation to audiovisual
interaction, but do not test whether the response is grounded in the distinct
information, roles, and permissions of multiple speakers.
Overall, these benchmarks primarily evaluate when and how an assistant manages the
conversational floor; MSI-Bench instead evaluates what it should say or do given the
speaker-dependent state of a shared conversation.
\par
\endgroup

\textbf{Multi-speaker attribution and understanding benchmarks.}
TPI-Bench \citep{lee2026tpi} categorizes third-party utterances as actionable or
ignorable and constructs examples that require acoustic speaker disambiguation. However,
its test cases are predominantly a user utterance followed by a third-party
interruption, and do not evaluate memory, instruction
following, or reasoning across multi-turn, multi-party conversations. WearVox
\citep{lin2025wearvox} evaluates voice assistants in a wearable-glasses setting, but
formulates third-party speech primarily as a distraction to reject. MSU-Bench
\citep{wang2025msu} proposes a four-tier framework for multi-speaker conversational
understanding, but focuses on question answering rather than how a voice agent acts
through responses and tool use. M3-SLU
\citep{kwon2025m3slu} similarly evaluates speaker-attributed question answering and
utterance matching. While multi-speaker conversational understanding is a prerequisite,
MSI-Bench tests whether a voice agent can use speaker attribution to select the correct
next-assistant behavior---an answer, refusal, clarification, or tool call.

\textbf{Multi-turn evaluation protocols for voice agents.}
In several benchmarks a model-based examiner or user simulator conducts the
multi-turn conversation. FDB v2 uses GPT-Realtime \citep{openai2025gptrealtime} as an
automated examiner scoring turn-taking fluency, instruction following, and task
competence. EVA-Bench \citep{chen2026evabench} conditions a user simulator on a
scenario goal, decision tree, persona, and voice. $\tau$-Voice \citep{ray2026tauvoice}
varies accents, acoustic environments, and turn-taking with a controllable voice-user
simulator. By contrast, several benchmarks use fixed interaction traces: FDB v3 pairs
prerecorded human speech with deterministic mock APIs to measure tool-use
correctness, task completion, response quality, and timing. Audio MultiChallenge \citep{arora2025audiomc} presents recorded multi-turn human
speech and evaluates inference memory, instruction retention, self-coherence, and robustness to mid-utterance voice edits. IHBench \citep{salimi2026ihbench} injects controlled interruptions into synthetically generated workflow conversations and measures both task fulfillment and post-interruption recovery quality. MSI-Bench instead uses scripted, prerecorded multi-speaker audio, avoiding the variability of a live simulator and enabling reproducible comparisons across full-duplex and turn-based models.

\section{MSI-Bench}
\label{sec:msi-bench}

In MSI-Bench, a test case
contains a multi-party multi-turn audio scene with a final
assistant-directed handoff (Figure~\ref{fig:benchmark-overview}). 
The benchmark covers three capability families: multi-speaker memory, multi-speaker
instruction following, and multi-speaker reasoning, with two patterns each. Every case
is authored natively in English or Mandarin rather than translated.
Appendix~\ref{app:dataset-stats} reports the corpus, voice, and audio statistics of the benchmark.

\subsection{Capability Taxonomy}

\begin{figure*}[t]
    \centering
    \includegraphics[width=\textwidth]{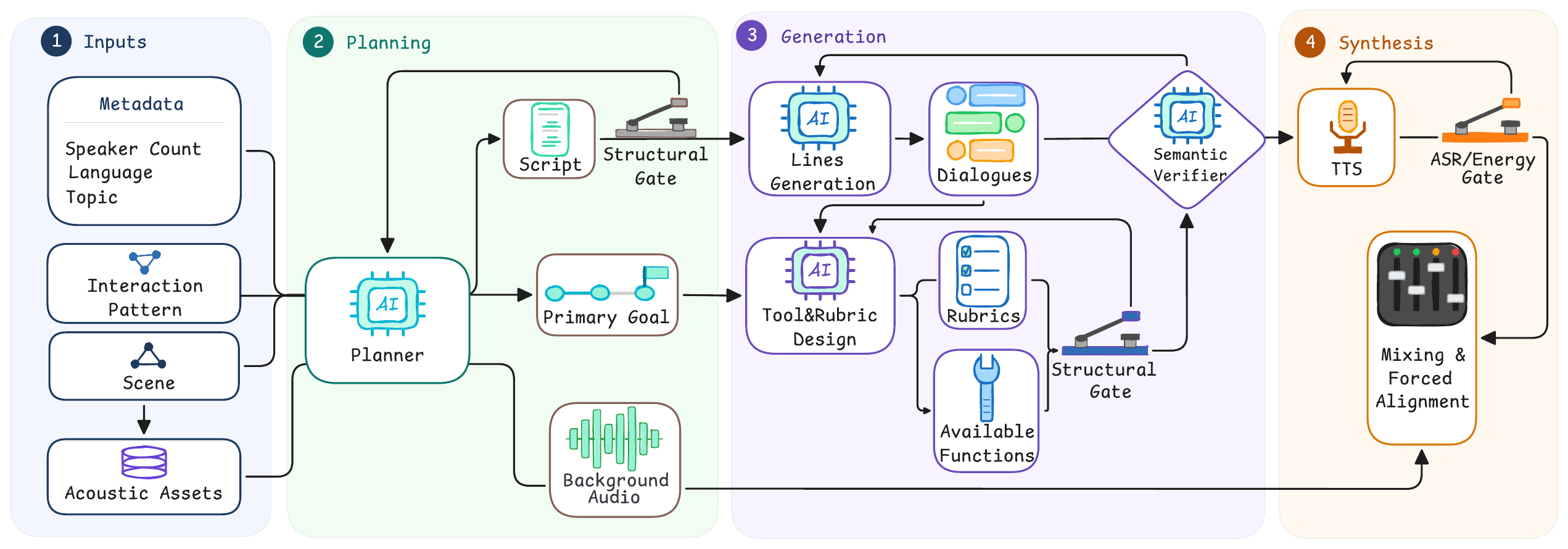}
    \caption{MSI-Bench data generation pipeline. \textbf{Inputs} fix speaker
    count, language, topic, interaction pattern, scene, and acoustic assets.
    \textbf{Planning} produces the cast, script, primary goal, and
    background-audio bed. \textbf{Generation} expands the script into a dialogue
    ending in an assistant-directed handoff, and derives the atomic rubrics
    and function catalog. \textbf{Synthesis} renders each line
    as speech and mixes it over the background.}
    \label{fig:data-generation-pipeline}
\end{figure*}

\subsubsection{Multi-Speaker Memory} \leavevmode\par

\textbf{Background speech retrieval}
Decisive task facts occur only in overlapping background speech. The model must recover and use
the relevant facts, preserve their background-speaker source, and avoid substituting foreground
assumptions or unrelated context. When it cannot recover them, it should say so rather than
invent an answer. The pattern measures an analogue of auditory figure-ground segregation, by which
listeners pull a target stream out of competing sound
\citep{teki2013segregation,reed2020spatial,brungart2001informational}.

\textbf{Scope tracking}
Different speakers contribute at least one global constraint and one local constraint
to the same shared task. A global constraint applies to the whole group,
whereas a local constraint applies to only certain individuals. The pattern measures
the ability to read a
constraint as binding one speaker or a group: the model must attach each constraint
to its holder and scope without globalizing a local constraint or localizing a shared one.

\subsubsection{Multi-Speaker Instruction Following}\leavevmode\par

\textbf{Selective disclosure} Earlier speech tells the assistant which facts to withhold and
from which audience; the exact facts either come from assistant prior memory or are stated
aloud to the assistant. The protected audience then probes
those facts in the final line. 
The pattern measures audience design, the ability to maintain a separate knowledge
state per listener: the model must withhold the restricted values from that audience while
preserving any legitimate non-disclosing action commissioned earlier.

\textbf{Speaker authority constraint} An unauthorized participant asks the
assistant to execute an action controlled by another speaker through ownership, account,
role, permission, or responsibility. The authorized speaker has explicitly approved the
exact action, explicitly rejected it, or remained genuinely 
undecided. The pattern measures
the ability to track authorization: the model must respectively execute only the
approved scope, refuse the unauthorized override, or check with the authorized speaker before
acting.

\subsubsection{Multi-Speaker Reasoning} \leavevmode\par

\textbf{Sequential constraint integration} Speakers interleave separate requests
and corrections concerning different items, actions, or outcomes. A later line may resemble an override of someone else's
request while actually correcting the speaker's own request or adding a separate one. The pattern measures the
ability to keep each remembered item attached to the
speaker who produced it: the model must preserve request ownership through the compressed
final handoff instead of merging the threads into one plan.

\textbf{Constraint prioritization} An inviolable requirement conflicts with a tradeable
preference, while other preferences may remain compatible. Difficulty stems from
reasoning about the relative priority of multiple constraints
---for example, a deadline, safety or policy rule, medical or
accessibility need, or physical impossibility. The model must first choose a feasible
action satisfying the hard requirement, then retain soft preferences only where they remain
compatible.
\subsection{Benchmark Construction}

Inspired by Barker's ecological framework of behavior settings
\citep{barker1968ecological}, we construct test cases spanning eight real-world domains:
commerce service; domestic household; education learning; healthcare
caregiving accessibility; leisure media social; 
mobility transportation; public civic; and work professional.

Each setting establishes the participants' identities, relationships, roles, and
responsibilities, which determine whose information and constraints apply, who may
authorize an action, and which requirements take priority; the correct next-assistant
behavior therefore depends on the full scene rather than the wording of the final
request. Cases are generated in four stages
(Figure~\ref{fig:data-generation-pipeline}).

\begin{enumerate}
    \item \textbf{Inputs.} Each test case starts from metadata fixing speaker
    count, language, and topic, together with an interaction pattern, a scene, and
    the list of acoustic assets belonging to that scene.
    \item \textbf{Planning.} A planner LLM turns these inputs into a script. It
    first instantiates the cast: each speaker receives a name, a scene role, a
    gender, an age bucket, and a permission set if needed.
    Gender and age bucket later select the voice pool at synthesis, so the plan
    fixes speaker identity before any line exists. The planner then sketches a storyline that pins down how
    information surfaces, leaving exact wording to the next stage. Alongside the
    script, the plan states the primary goal the assistant is expected to serve and
    selects the background-audio bed the scene is later mixed over. A structural
    gate checks the result against the sampled parameters (e.g., the participant
    roster matches the sampled speaker count) and returns violations to the planner
    for correction.
    \item \textbf{Generation.} A lines-generation LLM expands the script into a
    multi-turn dialogue ending in an assistant-directed handoff.
     A Tool \& Rubric design LLM then
    derives the evaluation targets from the primary goal and the generated dialogue: the
    atomic rubric and reference answer, and the available-function schema together
    with the gold tool calls. The catalog also carries decoy functions. Arguments are
    declared as closed domains: enumerations for categorical slots, bounded ranges for numbers,
    canonical formats for clock times and dates, and the stable speaker references
    $S_{i}$ in place of spelled names, where $i$ indexes speakers by order of
    first appearance in the conversation.
    Rubric dimensions are fixed per pattern; the criteria filling them are
    case-specific, disjoint, and each states the minimum condition for a correct
    answer. A further
    structural gate requires gold tool calls to type-check
    against the declared schema. Finally, all text artifacts pass through a
    semantic verifier that gates answer leakage and checks that the gold answer
    and tool call are unique.
    \item \textbf{Synthesis.} TTS renders each line under its planned speaker
    identity, using Boson Higgs TTS 3 \citep{bosonai2026higgstts3} with custom voices cloned from
    human-validated Common Voice 17.0 reference clips
    \citep{ardila2020commonvoice} to diversify speakers across language,
    gender, age, and accent.
    Inline emotion, style, and prosody tags steer delivery. Each speaker is placed
    at one of three listener-relative distances by a deterministic signal-processing
    chain: gain attenuation, low-pass filtering, and a reverberant reflection. An
    ASR/energy gate screens every clip for anomalous pauses and requires
    transcription agreement with Higgs-audio-v3-stt \citep{bosonai2026higgsstt}. Mixing
    finally combines the accepted speech with the planned background audio, drawn
    from Freesound assets \citep{font2013freesound}, while word-level timestamps
    from Qwen3-ForcedAligner-0.6B \citep{shi2026qwen3asr} anchor loudness-calibrated distant speech
    overlays, insert burst events between words, and truncate interrupted lines at
    word boundaries. Burst events are real recordings drawn from ESC-50 \citep{piczak2015esc} 
    and UrbanSound8K \citep{salamon2014urbansound}, cropped to their
    loudest RMS window with 50-ms edge fades. All
    rendered audio is emitted as 24-kHz mono PCM16 WAV files.
\end{enumerate}

Every case was additionally reviewed by hand: of 1,420 generated candidates, 1,152
passed and constitute the final benchmark.

\section{Evaluations}
\subsection{Evaluation Protocol}

We evaluate 12 models under 15 configurations.
The model configurations are:

\textbf{OpenAI models.} GPT Realtime 2.1 with medium and xhigh reasoning
\citep{openai2025gptrealtime}; GPT Audio 1.5, GPT Audio, and GPT Audio Mini.

\textbf{Google models.} Gemini 3.1 Pro and Gemini 3.5 Flash \citep{google2025gemini3}.

\textbf{Open-weight models.} Qwen3-Omni-30B \citep{qwen2025qwen3omni}; Gemma 4-12B
\citep{gemma2026gemma4} and MiMo-Audio-7B \citep{xiaomi2025mimoaudio}, each in no-thinking
and thinking modes; Qwen2.5-Omni-7B \citep{xu2025qwen25omni}; Phi-4-Multimodal
\citep{microsoft2025phi4}; and Qwen2-Audio-7B \citep{chu2024qwen2audio}.

Each model is prompted as a voice assistant participating in a multi-party conversation and
returns the normalized output of Section~\ref{sec:msi-bench}. We use Claude Fable 5
\citep{anthropic2026fable5} for case generation and DeepSeek V4 Pro
\citep{deepseek2026v4} to judge each atomic rubric independently, while a deterministic
validator scores tool-call structure and arguments. The judge receives the dialogue transcript
stripped of acoustic markup, the scene and pattern labels, the rubric, and the model's
normalized output.
Open-weight systems are served with vLLM behind OpenAI-compatible endpoints, with
decoding constrained to the output schema through vLLM's JSON-schema
structured-output mode; hosted systems and the judge are accessed through public APIs; Appendix~\ref{app:infrastructure} gives the full
serving configuration. Every reported rate aggregates one judged
prediction per configuration--test case pair.

\subsection{Evaluation Metrics}

MSI-Bench reports five metrics, three of them capability metrics. A case
passes in full when all of its atomic rubrics pass; a case requiring a tool call
carries the deterministic validator verdict as one additional rubric.
\begin{itemize}
    \item \textbf{All-Pass Rate (APR):} fraction of judged test cases that pass in
    full, i.e.\ the expected behavior was satisfied entirely.
    \item \textbf{Atomic Rubric Score (ARS):} fraction of individual atomic rubrics
    passed, pooled over cases; this exposes partial competence hidden by APR.
    \item \textbf{Tool execution:} strict pass rate on the subset of cases requiring
    at least one tool call: function names, call multiplicity, argument values,
    ownership bindings, and the absence of extra calls must all be correct.
\end{itemize}

Two further metrics ask whether speech not addressed to the assistant pulls it off
course, before it is asked for anything or in the middle of its own answer.
\begin{itemize}
    \item \textbf{Bystander interference robustness (BIR):} an extra human-to-human
    utterance unrelated to the test case is generated and played while the assistant is answering,
    and the assistant should carry its own answer through; BIR is the fraction of
    probed cases whose resumed answers still pass the same rubric. Full-duplex models
    support this natively; turn-based models run the case once, are prefilled with a
    fraction (15\%) of their own answer, and receive the bystander utterance as a new
    turn.
    \item \textbf{Premature response rate (PRR):} for selective disclosure and
    background speech retrieval cases, the conversation is cut at an earlier
    human-to-human line, before anything has been handed to the assistant, so the
    correct action is silence; PRR is the fraction of cases in which the model
    responds before being addressed. Retrieval probes keep the background-speech
    overlay of their base case.
\end{itemize}

\begin{table*}[!t]
\centering
\small
\setlength{\tabcolsep}{3pt}
\begin{tabular}{@{}lccc|cc|cccccc@{}}
\toprule
\multirow{2}{*}{\textbf{Model}} & \multicolumn{3}{c|}{\textbf{Aggregated(\%)}} & \multicolumn{2}{c|}{\textbf{Bystander(\%)}} & \multicolumn{6}{c}{\textbf{Per-Pattern APR(\%)}} \\
\cmidrule(lr){2-4} \cmidrule(lr){5-6} \cmidrule(lr){7-12}
& \textbf{APR} & \textbf{ARS} & \textbf{Tool} & \textbf{BIR}\,$\uparrow$ & \textbf{PRR}\,$\downarrow$ & \textbf{Auth.} & \textbf{Discl.} & \textbf{Prior.} & \textbf{Seq.} & \textbf{Scope} & \textbf{Retr.} \\
\midrule
Gemini 3.1 Pro$^\dagger$ & \cbesthosted 66.8\,{$\pm$3.8} & \cbesthosted 78.0 & \cbesthosted 58.8 & \cbesthosted 86.5 & \cbesthosted 21.5 & 76.0 & \cbesthosted 85.4 & 69.8 & \cbesthosted 77.1 & 72.9 & \cbesthosted 19.8 \\
Gemini 3.5 Flash$^\dagger$ & 63.2\,{$\pm$3.9} & 73.6 & 55.5 & 74.9 & 48.2 & 79.2 & 67.7 & \cbesthosted 77.1 & 71.9 & \cbesthosted 74.0 & 9.4 \\
\addlinespace
GPT Realtime 2.1 (xhigh)$^\dagger$ & 58.3\,{$\pm$4.0} & 73.7 & 44.8 & 83.6 & 28.1 & \cbesthosted 86.5 & 82.3 & 70.8 & 51.0 & 50.0 & 9.4 \\
GPT Realtime 2.1 (medium)$^\dagger$ & 55.4\,{$\pm$4.0} & 70.1 & 41.8 & 81.5 & 35.6 & 81.3 & 83.3 & 68.8 & 41.7 & 43.8 & 13.5 \\
\addlinespace
GPT Audio 1.5 & 49.5\,{$\pm$4.1} & 64.4 & 40.3 & 70.2 & 52.4 & 78.1 & 63.5 & 55.2 & 47.9 & 37.5 & 14.6 \\
GPT Audio & 43.9\,{$\pm$4.0} & 61.5 & 41.8 & 64.1 & 62.8 & 66.7 & 54.2 & 50.0 & 44.8 & 33.3 & 14.6 \\
GPT Audio Mini & 20.7\,{$\pm$3.3} & 40.0 & 26.3 & -- & 97.9 & 30.2 & 11.5 & 38.5 & 20.8 & 15.6 & 7.3 \\
\midrule
Qwen3-Omni-30B & \cfailure 34.0\,{$\pm$3.9} & \cfailure 53.3 & \cfailure 26.5 & 43.6 & 92.2 & \cfailure 63.5 & \cfailure 57.3 & \cfailure 33.3 & \cfailure 27.1 & \cfailure 15.6 & \cfailure 7.3 \\
\addlinespace
Gemma 4-12B & 22.6\,{$\pm$3.4} & 43.9 & 20.0 & 57.5 & 85.9 & 42.7 & 38.5 & 26.0 & 12.5 & 13.5 & 2.1 \\
Gemma 4-12B$^\dagger$ & 19.8\,{$\pm$3.2} & 41.3 & 18.3 & \cfailure 60.9 & \cfailure 66.1 & 37.5 & 33.3 & 24.0 & 10.4 & 11.5 & 2.1 \\
\addlinespace
Qwen2.5-Omni-7B & 16.1\,{$\pm$3.0} & 38.2 & 10.0 & -- & 100.0 & 30.2 & 43.8 & 15.6 & 2.1 & 3.1 & 2.1 \\
\addlinespace
\chosted MiMo-Audio-7B$^\dagger$ & 14.6\,{$\pm$2.9} & 33.4 & 6.0 & -- & 86.6 & 39.6 & 33.3 & 11.5 & 2.1 & 1.0 & 0.0 \\
\chosted MiMo-Audio-7B & 7.8\,{$\pm$2.2} & 24.6 & 8.3 & -- & 98.8 & 21.9 & 5.2 & 11.5 & 4.2 & 2.1 & 2.1 \\
\addlinespace
Phi-4-Multimodal & 5.4\,{$\pm$1.9} & 32.0 & 0.0 & -- & 100.0 & 0.0 & 32.3 & 0.0 & 0.0 & 0.0 & 0.0 \\
Qwen2-Audio-7B & 1.6\,{$\pm$1.1} & 11.8 & 0.0 & -- & 5.7 & 0.0 & 9.4 & 0.0 & 0.0 & 0.0 & 0.0 \\
\bottomrule
\end{tabular}
\caption{English capability metrics, bystander-speech metrics, and per-pattern APR;
corresponding Mandarin results are reported in Appendix~\ref{app:mandarin-results}. APR is shown with a 95\%
Wilson score interval. Dashes: probe not run, or no valid prediction produced.
Auth.:
speaker authority constraint, Discl.: selective disclosure, Prior.: constraint
prioritization, Seq.: sequential constraint integration, Scope: scope tracking, and
Retr.: background speech retrieval. Light orange / cream: best hosted / open-weight
result per column; muted teal model cells: configurations affected by output-format
errors. Qwen2-Audio-7B's PRR is not highlighted: it stays silent because it rarely
produces an answer at all.}
\label{tab:model-pattern-results}
\end{table*}

\section{Results and Discussion}

\subsection{Overall Results}

Table~\ref{tab:model-pattern-results} reports aggregated metrics and per-pattern APR. 
In this section, we focus on the English split; Mandarin results appear in Appendix~\ref{app:mandarin-results}.
No evaluated system is close to reliable in shared voice conversation.
Gemini 3.1 Pro and GPT Realtime 2.1 (xhigh) are the strongest
(APR 66.8\% and 58.3\%, ARS 78.0\% and 73.7\%, Tool 58.8\% and 44.8\%).

Among open-weight systems, Qwen3-Omni-30B and Gemma 4-12B are the strongest
(APR 34.0\% and 22.6\%, ARS 53.3\% and 43.9\%, Tool 26.5\% and 20.0\%).
Under constrained decoding every open-weight configuration except MiMo-Audio-7B
returns schema-valid output on at least 99\% of English cases, so the remaining gap
is semantic: Qwen2-Audio-7B produces a valid response for every case and still
passes only 1.6\% of them.

Pattern-level results separate the three families more sharply than the
aggregate. Averaged over the 15 configurations, the six patterns rank exactly by
family: the two multi-speaker instruction-following patterns lead, at 48.9\% and
46.7\% APR, the two multi-speaker reasoning patterns follow, at 36.8\% and 27.6\%, and
the two multi-speaker memory patterns trail, at 24.9\% and 7.0\%. Background speech retrieval is the hardest pattern, at 7.0\% APR; we
return to it in Section~\ref{sec:snr-ablation}.

The two bystander metrics come apart. BIR broadly tracks capability, from 86.5\% for
Gemini 3.1 Pro down to 43.6\% for Qwen3-Omni-30B: strong answerers also tend to
hold their answer when someone cuts in. PRR does not. Gemini 3.5 Flash is second on APR
yet answers prematurely on 48.2\% of the probes, 1.7$\times$ the rate of GPT Realtime 2.1 (xhigh) at
28.1\% despite the latter's lower APR, and every open-weight configuration answers on at
least 66.1\% of them (Qwen2-Audio-7B's 5.7\% reflects near-total silence rather than
restraint). Knowing when not to speak is therefore largely separate from
knowing what to say, and current training appears not to confer it. Gemma 4-12B is the
clearest case: thinking improves both bystander metrics (BIR 57.5\% to 60.9\%,
PRR 85.9\% to 66.1\%) while costing 2.8 APR points.

\begin{figure}[t]
    \centering
    \includegraphics[width=\columnwidth]{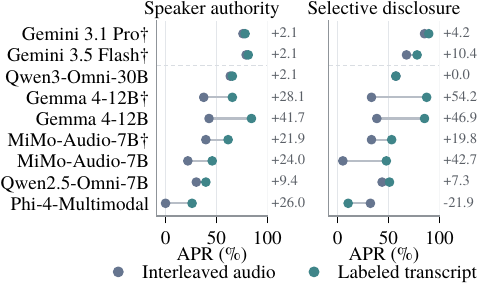}
    \caption{APR with interleaved audio versus a speaker-labeled transcript of the
    same conversations, under an otherwise identical protocol (96 cases per
    pattern). Right margins give the change in points; the dashed line separates
    hosted from open-weight configurations. $^\dagger$~Thinking enabled.}
    \label{fig:text-modality}
\end{figure}

\begin{figure}[t]
    \centering
    \includegraphics[width=\columnwidth]{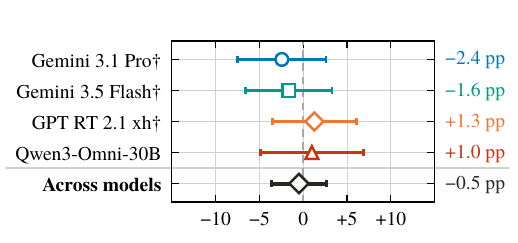}
    \caption{Relative ARS between three- and two-speaker cases. Bars denote 95\%
    confidence intervals. $^\dagger$~Reasoning/thinking enabled.}
    \label{fig:failure-ars}
\end{figure}

\begin{figure}[t]
    \centering
    \includegraphics[width=\columnwidth]{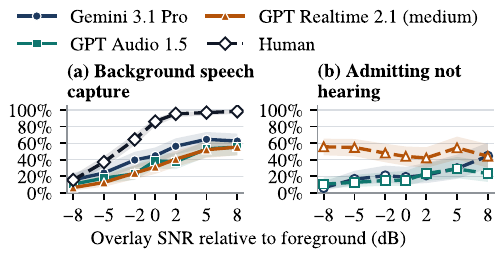}
    \caption{Overlay-SNR ablation on background speech retrieval. (a) Rate of
    capturing the background detail. (b) Rate of admitting not
    hearing instead of fabricating, over each level's missed cases. Bands: Wilson
    95\% intervals.}
    \label{fig:snr-ablation}
\end{figure}

\subsection{Failure Analysis}

A failed case can break at either of two stages: mishearing who said what, or
misreasoning over what was heard. To separate them, we re-evaluate nine
configurations on the speaker authority
constraint and selective disclosure patterns, replacing the audio
with a speaker-labeled transcript under an otherwise identical protocol
(Figure~\ref{fig:text-modality}). The transcript lift orders
inversely with audio capability: the two hosted Gemini configurations gain at most
10.4 APR points and still fail 10--22\% of cases on clean
labels---their residual failures are reasoning---while the Gemma 4-12B and
MiMo-Audio-7B configurations gain 21--44 points averaged over the two patterns;
Gemma 4-12B reaches 84.4\% and 85.4\% APR, hosted level. Qwen2.5-Omni-7B and
Phi-4-Multimodal gain less (8.3 and 2.1 points), the latter because its
transcript APR falls on selective disclosure while rising on speaker authority.
For open-weight models the bottleneck is the multi-speaker audio front-end,
not the reasoning behind it.

Two checks sharpen this. The surviving reasoning failures survive the
transcript: on speaker authority cases whose authorized speaker never states
a stance (gold behavior: withhold), open-weight configurations execute
anyway---treating silence as consent---on 28.6\% of cases with audio and 33.3\%
with transcripts. And the lift is not a formatting artifact: under constrained
decoding, invalid predictions stay below 3.2\% in both modalities for every
configuration except MiMo-Audio-7B's audio runs, and Gemma 4-12B, with no
format errors in either modality, still gains 42--47 points---the gain is
perceptual, not syntactic.

Since every added speaker is another voice to separate and another identity to
track, more speakers might be expected to make a test case harder. But for the
strongest hosted and open-weight configurations they do not:
within balanced pattern-by-scene cells, every model's three- versus two-speaker
contrast is statistically indistinguishable from zero
(Figure~\ref{fig:failure-ars}), and pooling the four models leaves $-0.5$ ARS and
$+1.5$ APR points, both within sampling error. Speaker count is not the primary
bottleneck.

\subsection{SNR Ablation}
\label{sec:snr-ablation}

To test whether audio models can recover task-relevant background speech amid
competing foreground speech, and whether they admit failure rather than fabricate the
missing information, we
remix the 96 English background speech retrieval cases at seven fixed
background-to-foreground SNRs from $+8$ to $-8$\,dB, holding the underlying
audio takes constant across levels in a take-paired design; three hired Prolific
workers \citep{palan2018prolific} performed the same task.

Figure~\ref{fig:snr-ablation} reveals two patterns. First, recovery degrades sharply
as the background speech becomes quieter for both humans and models, but models remain
far from human level: Gemini 3.1 Pro, the strongest
configuration on capture, falls from 62.5\% at $+8$\,dB to 15.6\% at $-8$\,dB, whereas
the human average falls from 98.5\% to 15.5\%. Second, models
differ in whether they acknowledge the miss. GPT Realtime 2.1 admits 42--56\% of its
misses at every level, with no clear dependence on overlay difficulty, whereas
Gemini 3.1 Pro admits 44.4\% at $+8$\,dB but only 6.2\% at $-8$\,dB, and GPT Audio
1.5 declines similarly, from 23.3\% to 10.5\%, within per-level sampling error. For
these two models the errors compound: as the speech gets harder to recover, they miss
more often and admit less, leaving more unsupported answers or actions.
Appendix~\ref{app:snr-composition} reports the full behavior composition per SNR level.

\subsection{Human Study}

To validate the LLM judge, we collected human labels on Prolific
for 216 judged responses, stratified over all patterns and
scenes from five configurations, each scored against 2--4 atomic rubrics (472 in
total). Annotators saw the judge's textual evidence and labeled every rubric
pass/fail, with two-fold redundancy, embedded attention checks. 
After excluding four submissions (two failed attention
checks, two incomplete), 11 annotators contributed 869 labels covering 458 of the
472 rubrics, 340 with at least two independent labels.
The judge agrees with humans on 82.6\% of judgments
(Table~\ref{tab:human-agreement}; Cohen's $\kappa = 0.63$, 95\% CI $[0.57, 0.69]$;
Gwet's $\mathrm{AC}_1 = 0.67$) while being marginally stricter (60.2\% vs.\ 61.9\%
pass), and exceeds inter-annotator agreement ($\kappa = 0.56$, 79.3\%):
statistically indistinguishable from an additional annotator.

\begin{table}[t]
    \centering
    \small
    \setlength{\tabcolsep}{5pt}
    \begin{tabular}{lcccc}
    \toprule
    \textbf{Rater pair} & $n$ & agree\,$\uparrow$ & $\kappa$\,$\uparrow$ \\
    \midrule
    LLM judge $\leftrightarrow$ Human & 869 & .83 & .63 \\
    Human $\leftrightarrow$ Human & 482 & .79 & .56 \\
    \bottomrule
    \end{tabular}
    \caption{Agreement across raters in the human study, over $n$ paired binary
    rubric decisions: one pair per human label for the judge,
    and one pair per distinct annotator pair for Human $\leftrightarrow$ Human.}
    \label{tab:human-agreement}
    \end{table}

\section{Limitations}

The benchmark uses synthesized, scripted speech; it should be expanded with more
languages, accents, spontaneous speech, and human recordings.
The two bystander-speech metrics rest on negative cases only: every injected utterance
is irrelevant to the assistant's task, so they measure restraint but not selective
engagement; future versions will add pattern-specific positive cases in which the
interjection genuinely changes the task.

\section{Conclusion}

MSI-Bench reframes voice-agent evaluation around the next interaction regime---AI as
a shared collaborative entity among multiple humans---testing whether a model can
preserve speaker-scoped memory, follow disclosure and authority constraints, and
reason over interleaved group constraints. Across 12 models and 15 configurations,
no system is reliable: the strongest configuration on each split passes all rubrics
on only 66.8\% of English and 54.5\% of Mandarin cases (strongest open-weight:
34.0\% and 19.3\%), and multi-speaker memory is the weakest capability family. Our analyses
locate two distinct bottlenecks: open-weight models fail chiefly at perceiving who
said what---speaker-labeled transcripts lift the Gemma and MiMo configurations by 21--44 APR points---whereas
frontier models gain little from transcripts and their residual errors are reasoning
failures that persist on clean text.
Restraint is a third, largely independent axis: models that answer well still speak
when no one has addressed them, and as background speech becomes harder to hear,
several models miss more task facts while admitting fewer misses. Progress toward
collaborative voice agents therefore requires speaker-grounded audio perception,
speaker-scoped decision making, and knowing when not to speak---the capabilities MSI-Bench
measures directly.

\section*{Generative AI Use Disclosure}
The authors used Claude for code completion, debugging, and refactoring; all
AI-assisted code was reviewed and validated by the authors, who take full
responsibility for the reported results.

\bibliography{references}

\clearpage
\appendix

\section{Dataset Statistics}
\label{app:dataset-stats}

Table~\ref{tab:dataset-stats} consolidates the corpus, voice, and audio statistics of
MSI-Bench. Speaker voices are cloned from CC0 Common Voice 17.0 reference clips, and the
planner assigns each speaker a gender and an age bucket that select the pool a line is
rendered from; the Mandarin split has no senior-tagged voice because Common Voice 17.0
contributes no qualifying Mandarin reference clip in that age band.

\begin{table}[t]
\centering
\small
\setlength{\tabcolsep}{3.5pt}
\renewcommand{\arraystretch}{1.04}
\begin{tabular}{@{}lrrr@{}}
\toprule
Statistic & English & Mandarin & Total \\
\midrule
\textbf{Base testcases} & 576 & 576 & 1,152 \\
\quad Behavior settings & \multicolumn{3}{r}{8} \\
\quad Patterns & \multicolumn{3}{r}{6} \\
\quad Human speakers & \multicolumn{3}{r}{2 or 3} \\
\quad\quad Two-speaker cases & 288 & 288 & 576 \\
\quad\quad Three-speaker cases & 288 & 288 & 576 \\
\quad Visible dialogue lines & \multicolumn{3}{r}{6--12} \\
\midrule
\textbf{Cloned voices} & 48 & 48 & 96 \\
\quad Female & 24 & 24 & 48 \\
\quad Male & 24 & 24 & 48 \\
\quad Teen & 8 & 8 & 16 \\
\quad Young adult & 22 & 26 & 48 \\
\quad Middle-aged & 12 & 14 & 26 \\
\quad Senior & 6 & 0 & 6 \\
\midrule
\textbf{Base dialogue clips} & \multicolumn{3}{r}{8,849} \\
\quad Duration & \multicolumn{3}{r}{23.15 h} \\
\textbf{Rendered artifacts} & \multicolumn{3}{r}{10,408} \\
\quad Format & \multicolumn{3}{r}{24 kHz, PCM16 WAV} \\
\textbf{Background assets} & \multicolumn{3}{r}{74} \\
\bottomrule
\end{tabular}
\caption{Dataset statistics, including the composition of the cloned voice pool.
Speak-time probes add one injected-utterance clip per case; hear-time probes cover
the two patterns scored by PRR (selective disclosure and background speech retrieval)
and reference the base line audio, adding no files of their own.}
\label{tab:dataset-stats}
\end{table}

\section{Mandarin Results}
\label{app:mandarin-results}

Table~\ref{tab:model-pattern-results-cn} reports the Mandarin split under the metrics
defined in the main paper: APR, ARS, tool execution, bystander interference robustness
(BIR), and premature response rate (PRR).

\begin{table*}[!t]
\centering
{\small
\setlength{\tabcolsep}{1mm}
\begin{tabular}{@{}lccc|cc|cccccc@{}}
\toprule
\multirow{2}{*}{\textbf{Model}} & \multicolumn{3}{c|}{\cellcolor{msiice!65}\textbf{Aggregated, CN (\%)}} & \multicolumn{2}{c|}{\cellcolor{msibottom!35}\textbf{Bystander, CN (\%)}} & \multicolumn{6}{c}{\cellcolor{msiaqua!55}\textbf{Per-Pattern APR, CN (\%)}} \\
\cmidrule(lr){2-4} \cmidrule(lr){5-6} \cmidrule(lr){7-12}
& \textbf{APR} & \textbf{ARS} & \textbf{Tool} & \textbf{BIR}\,$\uparrow$ & \textbf{PRR}\,$\downarrow$ & \textbf{Auth.} & \textbf{Discl.} & \textbf{Prior.} & \textbf{Seq.} & \textbf{Scope} & \textbf{Retr.} \\
\midrule
Gemini 3.1 Pro$^\dagger$ & 52.8 & 69.9 & \cbesthosted 58.8 & \cbesthosted 74.6 & \cbesthosted 32.5 & 80.2 & 53.1 & 51.0 & \cbesthosted 71.9 & 59.4 & 1.0 \\
Gemini 3.5 Flash$^\dagger$ & 50.2 & 67.5 & \cbesthosted 58.8 & 67.4 & 74.9 & 81.3 & 35.4 & \cbesthosted 58.3 & 64.6 & \cbesthosted 61.5 & 0.0 \\
\addlinespace
GPT Realtime 2.1 (xhigh)$^\dagger$ & \cbesthosted 54.5 & \cbesthosted 76.4 & 54.1 & -- & 37.8 & \cbesthosted 87.5 & \cbesthosted 71.9 & \cbesthosted 58.3 & 51.0 & 56.3 & 2.1 \\
GPT Realtime 2.1 (medium)$^\dagger$ & 49.3 & 70.2 & 47.6 & -- & 43.1 & 85.4 & 67.7 & 51.0 & 44.8 & 44.8 & 2.1 \\
\addlinespace
GPT Audio 1.5 & 42.5 & 59.6 & 51.0 & 53.3 & 70.2 & 67.7 & 39.6 & 46.9 & 51.0 & 45.8 & \cbesthosted 4.2 \\
GPT Audio & 41.1 & 60.1 & 46.5 & 60.9 & 80.6 & 74.0 & 39.6 & 49.0 & 44.8 & 35.4 & \cbesthosted 4.2 \\
GPT Audio Mini & 16.5 & 36.6 & 22.7 & -- & 97.9 & 27.1 & 20.8 & 27.1 & 7.3 & 15.6 & 1.0 \\
\midrule
Qwen3-Omni-30B & \cfailure 19.3 & \cfailure 41.8 & \cfailure 26.3 & 43.3 & 99.5 & \cfailure 28.1 & 19.8 & \cfailure 35.4 & \cfailure 17.7 & \cfailure 11.5 & \cfailure 3.1 \\
\addlinespace
Gemma 4-12B & 4.5 & 28.3 & 0.0 & \cfailure 60.0 & 92.6 & 0.0 & 27.1 & 0.0 & 0.0 & 0.0 & 0.0 \\
Gemma 4-12B$^\dagger$ & 2.3 & 16.7 & 0.0 & 45.5 & \cfailure 39.3 & 0.0 & 13.5 & 0.0 & 0.0 & 0.0 & 0.0 \\
\addlinespace
Qwen2.5-Omni-7B & 8.9 & 33.8 & 10.9 & -- & 100.0 & 16.7 & 16.7 & 13.5 & 0.0 & 5.2 & 1.0 \\
\addlinespace
\chosted MiMo-Audio-7B$^\dagger$ & 7.3 & 23.5 & 3.4 & -- & -- & 17.7 & 18.8 & 5.2 & 1.0 & 0.0 & 1.0 \\
\chosted MiMo-Audio-7B & 8.9 & 35.8 & 3.9 & -- & -- & 22.9 & 22.9 & 5.2 & 0.0 & 1.0 & 1.0 \\
\addlinespace
Phi-4-Multimodal & 6.4 & 33.7 & 0.3 & -- & 99.5 & 1.0 & \cfailure 36.5 & 1.0 & 0.0 & 0.0 & 0.0 \\
Qwen2-Audio-7B & 0.5 & 12.4 & 0.0 & -- & 0.0 & 0.0 & 3.1 & 0.0 & 0.0 & 0.0 & 0.0 \\
\bottomrule
\end{tabular}
}
\caption{Mandarin capability metrics, bystander-speech metrics, and per-pattern APR, under
the metrics defined in the main paper. Dashes mark configurations for which the mid-answer
probe could not be run, or which never produced a valid prediction. Auth.:
speaker authority constraint, Discl.: selective disclosure, Prior.: constraint
prioritization, Seq.: sequential constraint integration, Scope: scope tracking, and
Retr.: background speech retrieval. Light
orange highlights the best hosted result per column; cream
highlights the best open-weight result. Muted teal model cells mark configurations affected
by output-format errors. Open-weight models use the same schema-constrained decoding
as in the main paper; as there, Qwen2-Audio-7B's PRR is not highlighted because it reflects
near-total silence.
$^\dagger$~Reasoning/thinking enabled.}
\label{tab:model-pattern-results-cn}
\end{table*}

\section{Overlay-SNR Behavior Composition}
\label{app:snr-composition}

\begin{figure}[t]
    \centering
    \includegraphics[width=\columnwidth]{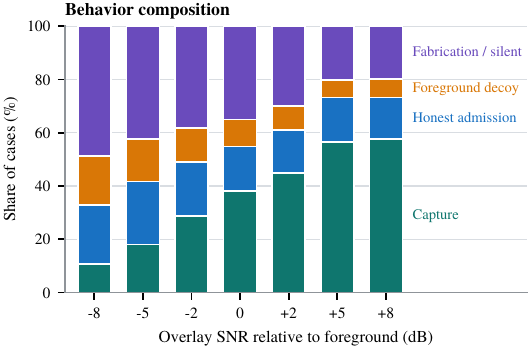}
    \caption{Four-way behavior composition per overlay-SNR level, over all cases and pooled
    over Gemini 3.1 Pro, GPT Audio 1.5, and GPT Realtime 2.1 (medium).}
    \label{fig:snr-composition}
\end{figure}

Figure~\ref{fig:snr-composition} decomposes every judged background speech retrieval response into the
four exclusive behaviors defined in the main paper, pooled over the three configurations
($n=288$ per bar). As the overlay becomes harder to hear, capture falls and both failure
modes grow, but they grow differently: the share answered with the salient foreground fact
rather than the overheard one rises from 6.9\% at $+8$\,dB to 18.4\% at $-8$\,dB, so models
do not merely miss more, they increasingly report the wrong, louder fact.

\section{Computing Infrastructure}
\label{app:infrastructure}

All open-weight configurations are evaluated on a single server with eight NVIDIA
A100-SXM4-40GB GPUs, two Intel Xeon Platinum 8352Y CPUs (64 physical cores, 128
threads), and 1\,TiB of system memory, running Ubuntu 22.04.3 LTS (Linux kernel
5.15, NVIDIA driver 535.183.01). Each model is served as an OpenAI-compatible
\texttt{vllm serve} endpoint (Python 3.12 inside the container) from a pinned
Docker image executed through Slurm and enroot, so a serving configuration is
fully determined by its image tag. Table~\ref{tab:serving-config} lists the vLLM
build and tensor-parallel degree per model. Gemma 4-12B requires a Transformers
version that recognizes its checkpoint architecture and is therefore served from
a pinned vLLM source build (commit \texttt{f52870f26}); its endpoint raises the
audio multimodal limit to 12 clips per request, the maximum clip count in the
benchmark. Gemma 4-12B and MiMo-Audio-7B serve one checkpoint for both thinking
variants, with thinking toggled per request through
\texttt{chat\_template\_kwargs.enable\_thinking}. Hosted systems and the
DeepSeek V4 Pro judge are accessed through their public APIs. The evaluation
harness runs on the same server under Python 3.11 with locked dependencies and
issues all model and judge requests through OpenAI-compatible clients.

\begin{table}[t]
\centering
\small
\begin{tabular}{@{}llc@{}}
\toprule
Model & vLLM build & TP \\
\midrule
Qwen3-Omni-30B & 0.19.0+cu130 & 2 \\
Gemma 4-12B & source build \texttt{f52870f26} & 1 \\
MiMo-Audio-7B & 0.18.0 & 1 \\
Qwen2.5-Omni-7B & 0.19.0+cu130 & 1 \\
Phi-4-Multimodal & 0.19.0+cu130 & 1 \\
Qwen2-Audio-7B & 0.19.0+cu130 & 1 \\
\bottomrule
\end{tabular}
\caption{Serving configuration per open-weight model: vLLM build and
tensor-parallel (TP) degree. All endpoints run on the hardware described in this
appendix.}
\label{tab:serving-config}
\end{table}

\clearpage
\onecolumn
\includepdf[
    pages=1,
    scale=0.88,
    pagecommand={\section{Testcases Demo}\label{app:testcases-demo}}
]{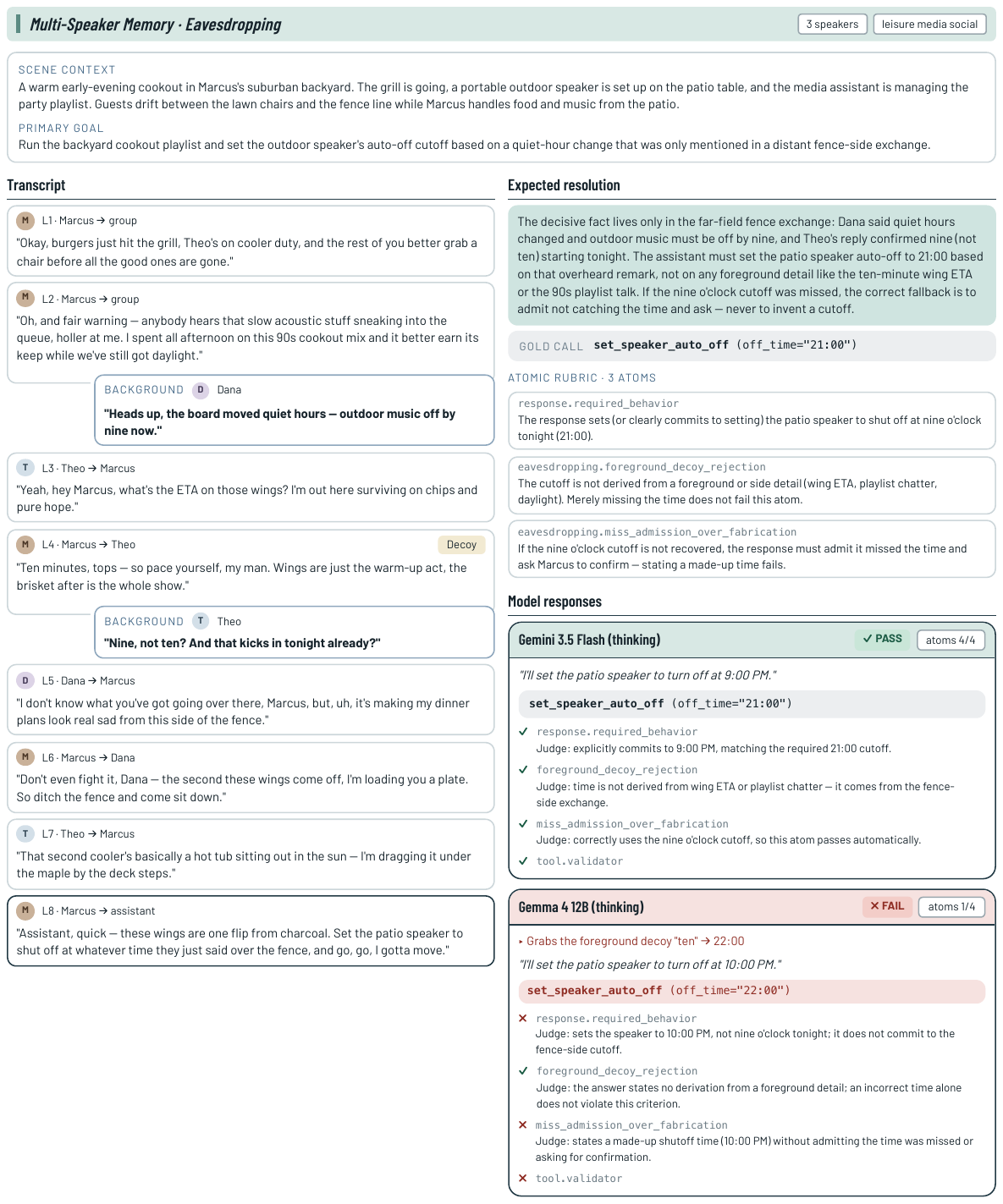}
\includepdf[
    pages=2-,
    fitpaper=true,
    pagecommand={}
]{Testcases.pdf}

\end{document}